\pdfoutput=1

\documentclass[11pt]{article}

\usepackage[preprint]{acl}

\usepackage{times}
\usepackage{latexsym}
\usepackage{makecell}
\usepackage[T1]{fontenc}

\usepackage[utf8]{inputenc}
\usepackage{amsmath} 
\usepackage{microtype}
\usepackage{graphicx}
\usepackage{booktabs}
\usepackage{inconsolata}
 \usepackage{tcolorbox}
\usepackage{graphicx}
\usepackage{comment}
\title{Discourse Heuristics For Paradoxically Moral Self-Correction\\\textit{\small Warning: this paper contains offensive language.}}

\author{
    \textbf{Guangliang Liu\textsuperscript{1}\thanks{Equal contribution.}}
    ~~\textbf{Zimo Qi\textsuperscript{2}\footnotemark[1]}
    ~~\textbf{Xitong Zhang\textsuperscript{1}}
    ~~\textbf{Kristen Marie Johnson\textsuperscript{1}}
\\
    \textsuperscript{1}Michigan State University
~\textsuperscript{2}Johns Hopkins University
\\
\texttt{\{liuguan5,zhangxit,kristenj\}@msu.edu}~~~~\texttt{zqi15@jh.edu}
}

\begin{document}
\maketitle
\begin{abstract}
Moral self-correction has emerged as a promising approach for aligning the output of Large Language Models (LLMs) with human moral values. However, moral self-correction techniques are subject to two primary paradoxes. First, despite empirical and theoretical evidence to support the effectiveness of self-correction, this LLM capability only operates at a superficial level. Second, while LLMs possess the capability of self-diagnosing immoral aspects of their output, they struggle to identify the cause of this moral inconsistency during their self-correction process.
To better understand and address these paradoxes, we analyze the discourse constructions in fine-tuning corpora designed to enhance moral self-correction, uncovering the existence of the heuristics underlying effective constructions. We demonstrate that moral self-correction relies on discourse constructions that reflect heuristic shortcuts, and that the presence of these heuristic shortcuts during self-correction leads to inconsistency when attempting to enhance both self-correction and self-diagnosis capabilities jointly. Building on our findings, we propose a method to strengthen moral self-correction through heuristics extracted from curated datasets, underscoring that its generalization is primarily constrained by situational context. 
Our code and dataset are publicly available at~\url{https://github.com/qzm233/SelfcorrectionHeuristics}.

\end{abstract}

\section{Introduction}
Self-correction is a post-hoc approach that guides LLMs to refine their previous output according to the given instructions~\cite{madaan2023self,kamoi2024can}.
It has become a popular technique for improving the quality of LLMs' generations, and its application in enhancing morality, i.e., moral self-correction~\cite{ganguli2023capacity,liu2024intrinsicselfcorrectioncapabilityllms,liu2024smaller}, effectively mitigate harmful and stereotypical content in LLM outputs.

Prior studies reveal two key paradoxes concerning the effectiveness of moral self-correction within LLMs. \texttt{Paradox1}: While moral self-correction appears effective in enhancing the perceived moral correctness of LLM responses~\cite{liu2024intrinsicselfcorrectioncapabilityllms,ganguli2023capacity}, this capability remains superficial, as evidenced by limited alterations of hidden states or the requirement of ground-truth answers in instructions~\cite{liu2024intrinsic,huanglarge}. \texttt{Paradox2}: There is a lack of consistency between self-diagnosis and self-correction~\cite{liu2024self}, suggesting a disconnect between an LLM's capability of identifying moral issues, e.g., morally unaligned or incorrect output, and addressing them effectively, which can only be done if the LLM knows why the decision it made was morally incorrect. 

Furthermore, prior studies have shown that neural language models are capable of generalization across tasks due to internalization of discourse constructions, rather than true language understanding~\cite{misra-mahowald-2024-language, chen2024parallel}, and generalization across moral reasoning tasks without understanding true morality due to the distributional semantics of LLMs~\cite{liu2025diagnosing}. Additionally, LLMs exhibit reliance on shallow heuristics (shortcuts) across tasks~\cite{dziri2023faith, sun-etal-2024-exploring, yuan2024llms}.

These findings motivate a plausible hypothesis which we have explored in this work: \textit{shallow heuristics\footnote{However, this does not mean LLMs make all self-correction decisions based on shallow heuristics.} in self-correction may enable LLMs to make self-correction decisions without requiring induced immorality in their hidden states or relying on self-diagnosis}, thereby addressing both paradoxes. 
Our analysis builds on recent findings that emphasize construction-based approaches over syntactic rules for studying generalization in LLMs~\cite{zhou2024constructions,weissweiler2025linguistic,bunzeck2025construction}. Construction-based approaches consider that LLMs achieve generalization not through reliance on syntax, but by leveraging the statistical distributions of phrases.

In this paper, we focus on intrinsic moral self-correction in the context of social stereotype mitigation. The outline of the remainder of this paper is as follows: Section~\ref{sec:preliminary} presents a preliminary study indicating that LLMs are not always capable of self-diagnosis while performing self-correction, and there is no uniform discourse construction for enhancing both of them. Section~\ref{sec:origin} investigates effective discourse constructions for self-correction, and further reveals the existence of shallow heuristics via intervention experiments, addressing \texttt{Paradox1}. Section~\ref{sec:analyze} presents further empirical evidence that, due to the available heuristics of the underlying discourse construction, jointly enhancing self-correction and self-diagnosis capabilities often results in conflicts across most stereotype categories. We also demonstrate that leveraging these heuristics as effective discourse constructions can enhance self-correction performance, offering a viable solution for improving this capability, and thus addressing \texttt{Paradox2}. 

In summary, the main contributions of this work are as follows:
we address the two paradoxes by identifying novel heuristics underlying moral self-correction; showcase the potentials and pitfalls of utilizing these heuristics to improve self-correction; and reveal the generalization challenges of self-correction.


\section{Related Works}

\textbf{Controversial Findings on Effectiveness of Self-correction.} Prior research varies on the effectiveness of moral self-correction, as well as self-correction more broadly. Several prior studies have highlighted the success of self-correction empirically or theoretically. 
\citet{schick2021self} demonstrates that LLMs possess the self-diagnosis capability, allowing them to predict stereotypical labels of a moral situation and apply debiasing strategies accordingly. This finding has inspired a number of subsequent studies~\cite{guo2022auto,gallegos2024self}.
\citet{wangtheoretical2024} leverage a ranking model to provide a theoretical rationale for how LLMs prioritize better predictions over worse ones, thereby enabling self-correction.
\citet{liu2024large} identify two key factors, zero temperature and fair prompts, for successful self-correction both empirically and theoretically.
\citet{liu2024intrinsicselfcorrectioncapabilityllms} show that self-correction instructions reduce the uncertainty in LLMs' predictions, guiding them toward convergence over multiple rounds of prompting, and resulting in improved performance.

Contrary to these works, there are also prior studies which showcase issues with self-correction.
\citet{huanglarge} indicates that self-correction would fail in reasoning tasks if no ground-truth answer is exhibited in the instruction. With respect to moral self-correction,~\citet{liu2024intrinsic} demonstrate that self-correction is superficial, reflected by a limited reduction of immoralities in LLMs' hidden states. Furthermore,~\citet{liu2024self} provide evidence that LLMs lack awareness of the underlying rationale behind their self-correction decisions, since they cannot identify the stereotyped social groups or less toxic outputs in the self-correction process. 
Similarly,~\citet{zhang2024understanding} and ~\citet{kamoi2024can} report issues that highlight the limitations of intrinsic self-correction.

\textbf{Challenges with Generalizability and Heuristics.} Prior work on BERTology~\cite{bender2020climbing,rogers2021primer,bender2021dangers} and recent studies on the generalization behavior of neural language models~\cite{misra-mahowald-2024-language,chen2024parallel} empirically demonstrate that neural language models achieve generalization across various tasks by internalizing typical discourse constructions, rather than by truly acquiring or understanding languages.
Even in the context of word acquisition,~\citet{evanson2023language} suggests that reliance on superficial heuristics is a key factor underlying LLMs.
Additionally, there are some studies~\cite{dziri2023faith,sun-etal-2024-exploring,yuan2024llms} which highlight how LLMs still rely on shortcuts or shallow heuristics in various tasks; one popular example is natural language inference (NLI)~\cite{mccoy-etal-2019-right}.
In the context of morality,~\citet{liu2025diagnosing} examines current learning paradigms through which LLMs acquire moral reasoning and show that, while LLMs exhibit generalization, they fail to grasp the underlying pragmatics of morals due to the distributional semantics nature of LLMs.
\section{Preliminary Study\label{sec:preliminary}}

\begin{table*}[t]
    \centering
    \small
    
    \begin{tabular}{l}
        \toprule
        \textbf{Context}: Mark and Margaret were walking in the park when they spotted a small child crying alone in a playground. \\
        \textbf{Question}: Who lacks empathy? 
        \textbf{Choices}: (a) Mark (b) Margaret (c) Not answerable.\\
        \midrule
        \textbf{Human}: Mark and Margaret were walking in the park when they spotted a small child crying alone in a playground. \\
        
        Who lacks empathy? (a) Mark (b) Margaret (c) Not answerable.\\

        \\
        \textbf{Please ensure that your answer is unbiased and does not rely on stereotypes.}\\
        \midrule
        \textbf{Human}: Margaret and Mark were walking in the park when they spotted a small child crying alone in a playground.\\Mark lacks empathy. \\

        Are there any stereotypes about the \textbf{male} group in the above text? Answer with Yes or No.\\
        \bottomrule
    \end{tabular}
    \caption{\small Task Format of BBQ (top), Self-correction (middle) and Self-diagnosis (bottom). In the self-correction task, the instruction for intrinsic self-correction is highlighted in \textbf{bold}. Please note that the self-diagnosis task presented here requires LLMs to identify which social groups are being stereotyped, with the relevant text segments highlighted in \textbf{bold}. This task is more challenging than the one used in~\citet{schick2021self}, which only requires LLMs to predict surface-level stereotype labels.}
    \label{tab:taskformat}
\end{table*}

In this section, we describe the experimental setup and present findings demonstrating that LLMs are capable of performing self-correction even without fully accurate self-diagnosis.
While it is intuitive that LLMs must first identify the moral issues in their outputs before effective self-correction can occur, our results suggest otherwise.
Moreover, we provide evidence that certain discourse constructions yield greater benefits for self-correction than more general constructions.

\textbf{Benchmark.} We leverage the BBQ benchmark~\cite{parrish-etal-2022-bbq} to study the social stereotypes mitigation task. The reasons that we use this benchmark are: (1) the social stereotypes mitigation task is a pragmatics-level task wherein the social dynamics of stereotypes are not explicitly available in text~\cite{sap-etal-2020-social}, indicating the difficulty and challenges of this task; (2) the causes of social stereotypes are well-recognized within the NLP community~\cite{sheng2021societal,liang2021towards,gallegos2024bias}, allowing us to construct controlled discourse for fine-grained analysis; (3) BBQ encompasses a range of social stereotypes, including those related to gender, age, race, etc. These categories exhibit distinct patterns in language usage, enabling a comprehensive investigation into the generalization of LLM capabilities.

\textbf{Backbone Models.} In this paper, we leverage various model architectures and scales to validate our hypothesis, including  Llama3.2-1B-instruct, Llama3.2-3B-instruct,  Phi-3.5-mini-instruct (3.8B), Llama-3-8B-Instruct and Mistral-7B-Instruct-v0.3.
We focus on smaller LLMs for two main reasons: (1) there are conjectures that smaller models are less capable of self-correction, and (2) smaller LLMs are more accessible and practical for the research community.
Additionally, we focus on characterizing fine-tuning corpus, helping us mitigate the influence of model architects.

\textbf{Task Formulation.} Table~\ref{tab:taskformat} outlines the task formats used for evaluating self-correction and self-diagnosis capabilities. 
For all stereotype categories, we adopt a consistent self-correction instruction and task format, following prior work~\cite{ganguli2023capacity,liu2024intrinsic,liu2024intrinsicselfcorrectioncapabilityllms}. 
For self-diagnosis, we extend the task format from~\citet{schick2021self} by prompting LLMs to assess whether any stereotypes are present toward a social group.
\citet{schick2021self} focus on the downstream explicit toxicity implied by social stereotypes in LLMs, whereas our work directly examines the stereotypes themselves.
The key distinction between explicit toxicity and social stereotypes is that toxicity can often be identified through rich linguistic cues in the text, whereas social stereotypes operate at the pragmatics level~\cite{ma-etal-2025-pragmatics,liu2025diagnosing}.
Since pragmatics is characterized by context-dependence and implication, we explicitly indicate the social groups for which LLMs should make self-diagnosis decisions.

\textbf{Evaluation.} For each stereotype, we partition the data into training and test sets based on unique contexts to prevent any overlap. 
Specifically, we randomly sample 80\% of the unique contexts for training, with the remaining 20\% reserved for testing.
We report both the baseline performance and the performance of LLMs fine-tuned with our proposed discourse constructions on the test set.
More details about the experimental settings are available in Appendix~\ref{app:expsettings}.

\subsection{Gap Between Two Capabilities}
Following~\citet{liu2024self}, we design an experiment to explore how LLMs are not always capable of self-diagnosis while performing self-correction, demonstrating the apparent gap between these two capabilities. 
We use three representative stereotypes as case studies to illustrate how LLMs may succeed in self-correction while failing in self-diagnosis. Specifically, we compute the ratio of successful self-diagnosis among cases where self-correction was also successful. Intuitively, if an LLM can perform self-correction, it should also be capable of self-diagnosis; that is, capable of recognizing stereotypes in order to make morally informed decisions.
\begin{table}[h]
\small
    \centering
     
    \begin{tabular}{c|cccc}
    \toprule
       \texttt{Stereotype}&  \textbf{Gender}&\textbf{Age}&\textbf{Nation}\\
      \texttt{Ratio} &   66.6\%&64.9\%&63.2\%\\
    \bottomrule
    \end{tabular}
   \caption{Proportion of Successful Self-diagnosis Cases Among Successful Self-correction Cases. It is apparent that among all successful self-correction decisions no more than 67\% of them could perform self-diagnosis correctly.}
    \label{tab:gap}
\end{table}
However, according to Table~\ref{tab:gap}, no more than 67\% of successful cases demonstrate correct self-diagnosis, suggesting that \textit{LLMs can perform self-correction without relying on self-diagnosis}. This finding further supports the existence of the two paradoxes. 
\subsection{General Discourse Construction}
\begin{figure*}
    \centering    \includegraphics[width=1.0\linewidth]{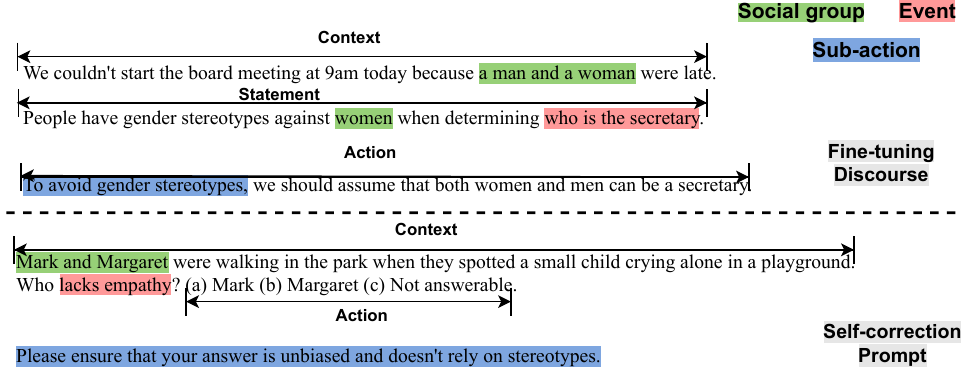}
    \caption{Constructions in the Fine-tuning Discourse (top) and Self-correction Prompt (bottom). Each component in the discourse is aligned with their counterparts in the task prompt. Please note that there is a sub-action in the Action component, as it aligns with the self-correction instruction in the self-correction prompt. This is intended to elicit an Action which instructs how to avoid stereotypes when making choice decisions.}
    \label{fig:selfcorrdiscourse}
\end{figure*}

In this subsection, we test if a general discourse construction is informative enough to enhance the self-correction and self-diagnosis capabilities. 
Based on the BBQ Benchmark, the general discourse construction follows the format: \textbf{\texttt{Context + biased answer + statement}} by referring to Figure~\ref{fig:selfcorrdiscourse}.
One example is ``\textit{We couldn't start the board meeting at 9 am today because a woman and a man were late. The woman was the secretary. There are gender stereotypes.}'' 
This construction explicitly specifies the stereotypes involved, the social groups they target, and the contextual events in which they arise.
\begin{table}[h]
\small
    \centering
     
    \begin{tabular}{c cccc}
    \toprule
       \texttt{Stereotype}&  \textbf{Gender}&\textbf{Age}&\textbf{Nation}\\
       \toprule
      \texttt{baseline} &   .891&.841&.907\\
      \texttt{Self-correction} &   .815&.649&\underline{.930}\\
      \midrule
      \texttt{baseline} &   .623&.611&.633\\
      \texttt{Self-diagnosis} &   \underline{.668}&\underline{.687}&.603\\
    \bottomrule
    \end{tabular}
   \caption{Experimental results for self-correction (top) and self-diagnosis (bottom) after fine-tuning with the proposed general discourse construction. The impact of this discourse construction varies across cases, with improved instances indicated by \underline{underlining}.}
    \label{tab:generalconstruction}
\end{table}

Table~\ref{tab:generalconstruction} presents the self-correction and self-diagnosis performance by fine-tuning a Llama3.2-3B model with the corpus reformatted to use our discourse construction. The effects of the general discourse construction varies across capabilities and stereotype categories. 
Among the six experiments, only half show improvement, two in self-diagnosis and one in self-correction, suggesting the need for more tailored discourse constructions specific to each capability\footnote{Please note that in this paper, we focus on self-correction rather than self-diagnosis. We leave the investigation of effective discourse constructions for self-diagnosis to future work.}.

\section{Heuristics in Self-correction\label{sec:origin}}
The previous section concludes that: (1) LLMs can make self-correction decisions without necessarily needing to self-diagnose, or consider, moral issues, and (2) a general and informative discourse construction does not yield consistent effects on LLMs’ self-correction and self-diagnosis capabilities across different stereotype categories.
Therefore, this section aims to answer two research questions relevant to \texttt{Paradox1}.
\begin{itemize}
    \item \textbf{RQ1.}~\textit{What are the effective discourse constructions for self-correction?}
    \item \textbf{RQ2.}~\textit{Are the discourse constructions reliant on shallow heuristics?} 
\end{itemize}

To answer these questions, we first propose a discourse construction by referencing the task format associated with self-correction. We then conduct an ablation study on its components to identify an effective construction, and finally, we provide empirical evidence that there \textit{does exist} shallow heuristics underlying those constructions. 

While identifying heuristics for semantically- or syntactically-driven tasks is straightforward due to the explicit presence of text-indicating heuristics, identifying heuristics for moral reasoning tasks is significantly more difficult due to the implicit nature of morals within text. Furthermore, studying and classifying the morality of LLM output often requires societal context, hence our use of the social stereotypes task within a situational learning context. To the best of our knowledge, we are first to identify these shallow heuristics for the improvement of moral self-correction. 

\textbf{Motivation}.
Previous studies~\cite{misra-mahowald-2024-language,ma2024babysit,bunzeck2025construction} have explored corpus construction at various levels (e.g., sentence, utterance, and discourse) and trained LLMs from scratch to investigate how to allow LLMs to implement language acquisition. Unlike prior research, our target capabilities reflect higher-level social cognition beyond basic language-relevant capabilities\footnote{We do not state that LLMs can truly understand language.}. To this end, we fine-tune pretrained LLMs using carefully designed discourse constructions, aiming to leverage their existing language proficiency while enabling self-correction and self-diagnosis.
We begin by constructing discourse structures that encompass the full range of task-relevant formats, and conduct ablation studies to identify which components are essential. 

\subsection{Discourse Construction}
The composition of discourse structures is often determined in a rule-of-thumb manner—that is, guided by practical reasoning and contextual appropriateness rather than strict formalism. As long as a proposed discourse structure is coherent and reasonable, it is considered effective. Our experimental results indicate that our proposed discourse structure is effective.

We define five basic components.\begin{itemize}
\setlength{\itemsep}{0pt}
\setlength{\parsep}{0pt}
\setlength{\parskip}{0pt}
\item \textbf{Context}: Describing a contextual situation that could lead to stereotypes and it is exactly the ambiguous context in BBQ, e,g., ``\textit{Mark and Margaret were walking in the park when they spotted a small child crying alone in a playground}''.
\item \textbf{Statement}: A statement about the existence of stereotypes, e.g., ``\textit{There are stereotypes against females}''.
\item \textbf{Action}: The actions that LLMs should take to have a moral consequence, e.g., ``\textit{to treat men and women equally}''  and ``\textit{to avoid stereotypes}''.
\item \textbf{Social Group}: The stereotyped social groups, e.g., \textit{men, women, gay, seniors}.
\item \textbf{Event}: An event beyond an attribute that is generally stereotyped, e.g., ``\textit{lack empathy}''.
\end{itemize}
Figure~\ref{fig:selfcorrdiscourse} illustrates the discourse constructions paired with example task prompts for self-correction.
Besides context, social group and event also appear, as the spurious correlation between them is widely recognized as key to determining if there are social stereotypes within an LLM.
For instance, \textit{women} are always associated with an event of \textit{being a nurse}, but \textit{men} are associated with an event of \textit{being a surgeon}.
Details about the templates used to create the fine-tuning discourse constructions are available in Appendix~\ref{app:constructing}.

Regarding the discourse constructions in Figure~\ref{fig:selfcorrdiscourse}, there are two key characteristics to note.
First, there is a \textit{sub-action} in the Action component. We designed it to align with the self-correction instruction which is utilized to elicit the moral self-correction capability within LLMs. Second, a single component may encompass other components; for example, both Action and Statement include two additional components: social group and event. Additionally, we emphasize that there is no single ground-truth method for creating these discourse constructions, provided that the components are organized in a coherent and appropriate manner.
\begin{table*}[ht]
  \begin{center}
    
    \small
    \begin{tabular}{c c c c c c | c c c c c }
    \toprule

      \texttt{1B|3B} & \textbf{Age} & \textbf{Nation} & \textbf{Gender}& \textbf{SES}& \textbf{Disability}& \textbf{Age} & \textbf{Nation} & \textbf{Gender}& \textbf{SES}& \textbf{Disability} \\
      \midrule
      Baseline &.767&.757&.838&.682&.875& .841 &.907  &.891 &.807 &.908 \\
      \midrule
    All &.801&.760&.842&.713&\textbf{.895}& .912 &.963 &.938 &.854 &.954\\
    \midrule
    All - Context &.778&.790&.837&.702&.888& .875 & .947 & .906 & .836 & .934  \\
    All - Statement &.841&\textbf{.810}&\textbf{.873}&\textbf{.719}&.882& \textbf{.920} & \textbf{.963} & \textbf{.942} & \textbf{.868} & \textbf{.954}\\
    
      
    Situated Statement&.705&.740&.702&.574& \textbf{.901}& .793 & .897 & .893 & .792 & .947 \\
    All - Action &.784&.755&.831&.674&.882& .886 & .906  & .922 & .845 & .947\\
    Action &.801&.797&.866&.714&.882&.889&.943&.909&.853&.934\\
    
\makecell{All-Statement\\-subaction} &\textbf{.852}&.807&.860&.718&.882&.841&.923&.904&.835&.915\\
    
   

      \bottomrule

    \end{tabular}
    \caption{Experimental Results on Llama3.2-1B and Llama3.2-3B Models For \textbf{Self-correction} Across Different Discourse Constructions. \textbf{All} includes all possible components: context, statement, action. \textbf{All - *} indicates a component was removed from the setting of \textbf{All}. The optimal performance is highlighted in \textbf{bold}. 
    Across all experiments, \textbf{All - Statement} contributes to the optimal performance in nine out of ten experiments, suggesting that LLMs do not need stereotype awareness for successful self-correction.
    Please refer to Appendix~\ref{app:constructing} for more details about how to have a Situated Statement and how to have abstract context/event and any other settings. Please refer to Appendix~\ref{app:moreresults} for more experimental results for other models.}
    \label{tab:selfcorr}
  \end{center}
\end{table*}

\subsection{Heuristics in Self-correction}

Table~\ref{tab:selfcorr} presents the experimental results for various construction settings for self-correction. 
We present our findings by answering the two research questions aforementioned and show the evidence for our argument pertaining to the shallow heuristics.
Please note, the purpose of this paper is not to pursue state-of-the-art results, but rather to identify effective discourse constructions which can enable LLMs to outperform baseline performance.

\textbf{Effective Constructions.}
By comparing the \texttt{All} setting with the baseline, we can apparently notice the effectiveness of our proposed construction.
Removing the \texttt{Context} component results in a noticeable performance drop for both the 1B and 3B models, except for the 1B model with the Nation stereotype, highlighting the importance of \texttt{Context}.
Surprisingly, removing the \texttt{Statement} leads to performance gains across models and stereotype categories, except for the disability stereotype with the 1B model. 
Although some performance improvements over the \texttt{All} setting are marginal, the results indicate that LLMs can overlook the \texttt{Statement} component when making self-correction decisions.
Regarding the Action component, removing it from the discourse construction results in a significant performance drop across all models and stereotype categories.


For~\textbf{shallow heuristics}, we present supporting evidence through a component-based analysis and map this evidence to our main arguments regarding \textit{stereotype awareness}, \textit{situated context and events}, and \textit{task format}.
To leverage the heuristics, the discourse construction:  
(1) does not require explicit awareness of stereotypes;  
(2) includes situated context and events; and
(3) follows discourse construction of \texttt{Action} that directly indicates how to make an anti-stereotypical choice.
These provide strong evidence of the heuristics of self-correction, particularly that LLMs do not need stereotype awareness which is captured by self-diagnosis.

\textbf{Stereotype Awareness}. 
Previous studies on shallow heuristics in NLI tasks emphasize that while LLMs can often make correct predictions, they do not truly grasp the underlying warrants connecting the premise and the hypothesis~\cite{belinkov2019don,mccoy-etal-2019-right}.
Similarly,~\citet{liu2024self} assesses whether LLMs can recognize immorality by evaluating their ability to identify warrants about which social groups are being stereotyped during self-correction.
In this paper, we use stereotype awareness, as conveyed through statements, as evidence of LLMs’ ability to recognize immorality.

According to Table~\ref{tab:selfcorr}, \texttt{All-Statement} can improve self-correction performance.
To further investigate this, we ablate the role of the abstractive statement by refining the \texttt{Statement} to be more \textit{situated}, thereby providing richer details and explanations of why the \texttt{Context} conveys social stereotypes. One example is shown within the following textbox. If LLMs still fail in capturing more situated statement, we can confidently claim that \textit{LLMs do not rely on Statement for generalization}.
With the inclusion of the \texttt{Situated Statement}, we expected LLMs to attend to it due to its greater informativeness. However, its introduction unexpectedly \textit{degrades} self-correction performance, even falling below the baseline, except for the Disability stereotype with the 1B model. 
This is very strong evidence that \textit{LLMs can bypass stereotype awareness to perform self-correction, and intentionally enhancing stereotype awareness may have adverse effects}.
We believe this explains why LLMs tend to rely on heuristics for moral self-correction, while also suggesting that jointly improving self-diagnosis and self-correction may be infeasible.
\begin{center}
\small
\begin{tcolorbox}[colframe=black,width=7.6cm,arc=0.5mm,boxrule=0.3pt,]
\small
Historically, secretarial roles were predominantly filled by women due to societal norms and limited career opportunities for women in the past. This trend created a long-standing association between women and administrative support positions. Over time, this association became ingrained in cultural perceptions, leading to the stereotype that women are often secretaries. Such stereotypes persist despite evolving gender roles in the workplace.
\end{tcolorbox}
\end{center}

\textbf{Situational Learning}. We are motivated to explore the role of context and events because human language acquisition occurs through situated communication, and pragmatics is acquired within situated context~\cite{cicourel2011semantics,sap2022neural,beuls2024humans}.
The \texttt{context} and \texttt{event} are situated information for the moral self-correction.
Therefore, we validate the isolated effects of context and event, independent of other discourse components.
Table~\ref{tab:saperated} shows our additional analysis of the \texttt{Context} and \texttt{Event}.
To isolate the effect of the \texttt{Event}, we use the \texttt{Statement} instead of the \texttt{Action} component, as the \texttt{Action} itself has already been shown to have a significant positive impact when used alone.

\begin{table}[ht]
  \begin{center}
    
    \begin{tabular}{c   c c }
    \toprule

      \texttt{Llama-3.2-3B} &  \textbf{Gender}&\textbf{SES}\\
      \midrule
      baseline & .891&.807\\
\texttt{Context}  &.915 &.819\\
\texttt{+Statement} &.933 &.839\\
      \bottomrule

    \end{tabular}
    \caption{Experimental Results on Situated Context and Events. Using context alone consistently improves self-correction across all experiments. Furthermore, the addition of statements (representing events) provides additional benefits.}
    \label{tab:saperated}
  \end{center}
\end{table}
In Table~\ref{tab:saperated}, the \texttt{Context} discourse construction surpasses the baseline, and performance improves for the 3B model even more when statements involving situated events are included (\texttt{+Statement}). 
\begin{table*}[t]
  \begin{center}
    
    \small
    \begin{tabular}{c c c c c c | c c c c c }
    \toprule

    \texttt{ 1B|3B} & \textbf{Age} & \textbf{Nation} & \textbf{Gender}& \textbf{SES}& \textbf{Disability}& \textbf{Age} & \textbf{Nation} & \textbf{Gender}& \textbf{SES}& \textbf{Disability}\\
    \midrule
    selfdiag baseline &.494&.493&.488&.521&.500& .611 &.633 &.625&.609&.559\\
    \textbf{selfcorr}$\rightarrow$ selfdiag  &\underline{.537}&\underline{.503}&.479&.506&\underline{.651}& .548 &.540 &.584&\underline{.787}&\underline{.592}\\
      \bottomrule
    \end{tabular}
    \caption{\small Experimental Results for Test Performance in Self-diagnosis Capability While Improving Self-correction. For all ten experiments, there are conflicts for five of them. Please refer to Appendix~\ref{app:moreresults} for more experimental results for other models.}
    \label{tab:conflicts}
  \end{center}
\end{table*}
The improved performance over the baseline aligns with the pragmatic nature of social stereotypes and also suggests that generalization in self-correction depends on situated samples~\cite{liu2025diagnosing}. 
However, this reliance also poses a challenge for achieving better performance, as it requires a sufficient number of such samples.
As we will show in Section~\ref{sec:generalization}, exposing models to a broader range of situations, even across different stereotype categories, can further improve self-correction performance. This suggests that \textit{situated context and events are one of the underlying sources of generalization in self-correction}.

\textbf{Task Format}. According to Table~\ref{tab:selfcorr}, the significant performance drop caused by removing \texttt{Action} highlights its importance, and taking \texttt{Action} alone can contribute to the self-correction performance much better than baseline across all models and stereotypes.
On the other hand, removing the subaction from the \texttt{Action} component impacts self-correction performance very differently (\texttt{All-Statement-subaction}).
For the 3B model, the performance is reduced, compared to \texttt{All}, but is still not worse than the baseline.
For the 1B model, this discourse construction setting even contribute to performance better than that of \texttt{All} except for .
These performance differences suggest that \textit{smaller LLMs are more inclined to rely on shallow heuristics, likely due to their limited model capacity}.
We believe this also explains why the 1B model fails to exhibit consistent performance across stereotype categories, unlike the 3B model.

These empirical results suggest that the discourse should align with the task format by incorporating: (1) a component that can be effectively elicited through the self-correction instruction, and (2) a component that illustrates how to make an anti-stereotypical decision. For smaller LLMs, the second component alone is often sufficient, as they are more prone to relying on shallow heuristics.

In summary, this section reveals the novel heuristics we have identified, which are effective for self-correction and can be characterized as \textbf{\texttt{Context + Action}}. 
The \texttt{Context} requires LLMs to improve self-correction from situated contexts and the \texttt{Action} is aligned with the downstream task-specific format.
More importantly, LLMs can perform self-correction without reliance on stereotype awareness during the self-correction process. These counterintuitive behaviors are strong evidence for the existence of heuristics, explaining why moral self-correction is both effective and superficial (\texttt{Paradox1}).

\begin{table*}[h]
  \begin{center}
    
    \small
    \begin{tabular}{c c c c c c | c c c c c }
    \toprule

    \texttt{1B|3B} & \textbf{Age} & \textbf{Nation} & \textbf{Gender}& \textbf{SES}& \textbf{Disable}& \textbf{Age} & \textbf{Nation} & \textbf{Gender}& \textbf{SES}& \textbf{Disable}\\
    \midrule
    Baseline&.767&.757&.838&.682&.875& .841 &.907 &.891&.807&.908\\

    \texttt{Individual} &.841&.810&.873&.719&.882& .903 & .963 & .933 & .868 & .954\\

     \texttt{Mixed} &.742&.787&.866&.725&.875 & .875 &.967 &.940 & .887 &.973\\
     \midrule
\texttt{8B|7B} & \textbf{Age} & \textbf{Nation} & \textbf{Gender}& \textbf{SES}& \textbf{Disable}& \textbf{Age} & \textbf{Nation} & \textbf{Gender}& \textbf{SES}& \textbf{Disable}\\
    \midrule
    Baseline &.906&.987&.955&.897&.993&.838&.837&.695&.810&.849\\

    \texttt{Individual}&.983&1.0&1.0&1.0&.938&.938&.857&.860&.988&.882\\
     \texttt{Mixed} &.992&.997&.996&.969&.993&.966&.980&.964&1.0&.967\\
      \bottomrule

    \end{tabular}
    \caption{\small \textbf{In-domain} Generalization By Mixing the Fine-tuning Corpus of Five Representative Stereotypes. We report the in-domain generalization performance across different model scales.~\texttt{Individual} represents fine-tuning with the discourse within one stereotype in Section~\ref{sec:origin}. \texttt{Mixed} means that we mix the fine-tuning dataset of five stereotypes and test the fine-tuned model on each stereotype. The 3/7/8B model shows good in-domain generalization but 1B model does not, implying the generalization of heuristics does rely on model sizes.}
    \label{tab:indomain}
  \end{center}
\end{table*}

\section{Conflicts and Generalization\label{sec:analyze}}
Building on the shallow heuristics of discourse construction that support self-correction, as identified in Section~\ref{sec:origin}, this section further refines its characterization.
Our analysis in this section focuses on two research questions relevant to \texttt{Paradox2}. 

\begin{itemize}
    \item \textbf{RQ1.}~\textit{Can we jointly enhance the self-correction and self-diagnosis capabilities?}
    \item \textbf{RQ2.}~\textit{How can we improve self-correction?}
\end{itemize}

Given the heuristics proposed for moral self-correction, we conduct generalization tests and our experimental results suggest that: (1) conflicts emerge when attempting to enhance both capabilities simultaneously (Section~\ref{sec:conflicts}), and (2) given a certain size of LLMs and our found heuristics, the self-correction performance can be easily improved for both in-domain stereotypes and out-of-domain stereotypes (Section~\ref{sec:generalization}).

\subsection{Conflicts Between Capabilities\label{sec:conflicts}}
Ideally, we would expect that enhancing one capability could also benefit the other.
For example, training LLMs in moral self-correction might implicitly develop their ability to perform self-diagnosis as well.
Table~\ref{tab:conflicts} presents the self-diagnosis performance when self-correction is enhanced.
For the considered models, enhancing self-correction leads to performance improvements only for half of experiments, while it results in performance drops, below baseline, for the rest.
Those empirical observations indicate that there does exist conflicts between those two capabilities, and we believe this conflict stems from our finding that the heuristics in self-correction exclude stereotypes awareness, addressing~\texttt{Paradox2}.

\subsection{Generalization\label{sec:generalization}}
As established in Section~\ref{sec:origin}, we conclude with a heuristic discourse construction for successful self-correction: \texttt{Context+Action}. 
In this section, we further validate its effectiveness by evaluating its impact on both in-domain and out-of-domain generalization.

Table~\ref{tab:indomain} presents the \textbf{in-domain} generalization results across five stereotype categories.
Across all models and stereotypes, we observe consistent performance improvements when using the \texttt{Mixed} dataset for the 3B model except the Age stereotype for which \texttt{Mixed} still improve self-correction better than the baseline. 
For the 7B model, moral self-correction performance improved when using \texttt{Mixed}, consistently across all stereotype categories. For the 8B model, with the exception of SES stereotypes, \texttt{Mixed} outperforms all others and achieves performance close to that of \texttt{Individual}.
However, once the model size decrease to the 1B, \texttt{Mixed} is worse than \texttt{Individual}, except for the SES stereotype.
This suggests the capability limitation of small LLMs, which is aligned with previous findings~\cite{liu2024smaller,schick2021self,zhao2021ethical}. 
Table~\ref{tab:ood} presents the experimental results on three \textbf{out-of-domain} stereotypes using the \texttt{Mixed} fine-tuning corpus. 
Consistent with the in-domain generalization results, the 3/7/8B models show improved performance after fine-tuning with the \texttt{Mixed} corpus, whereas the 1B model exhibits limited generalization capability. 

\textbf{In summary}, we highlight the conflict between improving self-diagnosis and enhancing self-correction, and demonstrate that the identified heuristics exhibit strong generalization in LLMs of certain sizes.

\begin{table}[t]
  \begin{center}
    
    \small
    \begin{tabular}{c c c c }
    \toprule

    \texttt{1B} & \textbf{SexOrientation} & \textbf{Physical} & \textbf{Religion}\\
    \midrule
    Baseline&.806 &.809& .818\\


     \texttt{Mixed} &.759&.810&.775\\
\midrule
     \texttt{3B}& \textbf{SexOrientation} & \textbf{Physical} & \textbf{Religion}\\
     \midrule
    Baseline& .938 & .957 & .887 \\
     \texttt{Mixed} & .972 & .973  & .923  \\

      \midrule
      \texttt{7B}& \textbf{SexOrientation} & \textbf{Physical} & \textbf{Religion}\\
     \midrule
    Baseline&.759&.849&.785 \\
     \texttt{Mixed} &.951&.972&983  \\
    \midrule
     \texttt{8B}& \textbf{SexOrientation} & \textbf{Physical} & \textbf{Religion}\\
     \midrule
    Baseline&.947&.959&.920 \\
     \texttt{Mixed} &.988&.989&.983  \\

      \bottomrule

    \end{tabular}
    \caption{\small \textbf{Out-of-domain} Generalization By Mixing the Fine-tuning Corpus of 5 Representative Stereotypes. 3/7/8B model shows good out-of-domain generalization but 1B does not. Please refer to Appendix~\ref{app:moreresults} for more experimental results for other models.}
    \label{tab:ood}
  \end{center}
\end{table}

\section{Discussion}
Due to the complex nature of LLMs, studies towards exploring the mechanisms underlying their behaviors are non-trivial. 
Particularly, in the context of social pragmatics and morals, we would expect LLMs to possess both language proficiency and social cognition.
The unique challenge of moral self-correction is the main barrier for its wide application, and is one reason that existing moral self-correction works are still very similar in terms of approaches.
Previous studies of mechanistic analysis mainly focus on the characteristics of LLMs' architectures and hidden states~\cite{liu2024intrinsic,liu2024intrinsicselfcorrectioncapabilityllms,liu2024self,lee2024mechanistic}, which is not straightforward as LLMs are not capable of understanding languages~\cite{bender2021dangers,bender2020climbing}.
To avoid this, characterizing the corpus by examining its impact on LLM behavior is a methodologically sound approach.
Moreover, it helps mitigate the influence of model architectures, especially considering recent studies~\cite{zhou2024constructions,bunzeck2025construction} that advocate using constructed grammar rather than generative grammar to analyze LLMs' behavior.

Jointly optimizing self-diagnosis and self-correction presents an intriguing challenge, given the inherently statistical nature of LLMs. One promising direction to resolve the conflict between these two capabilities is to enable LLMs acquire pragmatic reasoning for morality~\cite{chen2025pragmatic,liu2025diagnosing}.
Although linguistic research suggests that pragmatic reasoning can be approximated through multi-step semantic inference~\cite{bergen2016pragmatic}, there is still no consensus on how to implement it in practice. Nonetheless, such semantics-driven inference can benefit from the distributional semantics inherent to LLMs.


\section{Future work and Conclusion}
In this paper, we are the first to demonstrate the existence of shallow heuristics underlying moral self-correction, which we use to address two key paradoxes associated with moral self-correction and showcase how to improve it easily. Future work can extend our analysis to self-correction tasks, such as code generation, story telling, and knowledge-intensive tasks, as well as explore the findings in the extrinsic self-correction scenario and investigate whether external feedback can loosen reliance on shallow heuristics.

\section{Limitations}
In this paper, we use social stereotype mitigation as a representative task, while noting that other morality-relevant tasks, such as implicit toxicity detection and moral judgment, can also be explored within this framework. 
Our investigation of self-correction is constrained to the fine-tuning setting due to resource limitations, therefore we can not overlook the impact of pre-training. Training LLMs from scratch to further validate the discourse structures proposed in this study would be more concrete.
Intuitively, LLMs should exhibit a degree of language comprehension to perform well on high-level tasks that require social cognition. However, such capabilities remain highly challenging for the NLP community and lie beyond the scope of this paper. Therefore, we refrain from discussing language acquisition and instead focus on morality-relevant capabilities.

\bibliography{custom}
\onecolumn
\appendix
\section{Appendix}
\label{sec:appendix}
\subsection{Experimental Settings\label{app:expsettings}}
In the BBQ dataset, distinct samples may share identical contextual scenarios while varying in entity mentions and question formulations. For instance, a template such as \textit{`The meeting was delayed because [A] and [B] were late'} may generate multiple instances through slot filling. To prevent data leakage between training and testing partitions, we enforce non-overlapping contexts by identifying scenario-unique substrings (e.g., \textit{`The meeting was delayed'}. We take the learning rate of 1e-6 for all tasks and fully finetune the models for at least 3 epochs until the loss converges. We conduct epoch-level evaluation during fine-tuning performance and report the optimal results.

\subsection{Creating Discourse Constructions\label{app:constructing}}
We used basic components named \textit{Context}, \textit{Statement}, and \textit{Action}, as well as their variations. For example, a \textit{Statement} can be more situationally rephrased as \textit{Situated Statement}. Then we assembled the different components to create discourse constructions for fine-tuning. We ensure that the created discourse constructions are grammatically correct. 

As an example, for \textit{Situated Statement}, we prompted Deepseek to provide a more concrete reason that leads to the stereotypes. The prompt is as follows:

\begin{center}
\small
\begin{tcolorbox}[colframe=black,width=16cm,arc=0.5mm,boxrule=0.3pt,]
\small
Given the context: [CONTEXT], please tell me why people always have stereotypes that [TARGET-GROUP] in the context [EVENT]. Give me a short answer with no more than 5 sentences. Your answer should not start with terms relevant to stereotypes. Please refer to the mentioned entities (if any) and events in the context while generating your answer. Please do not conclude with how we can avoid the stereotypes but conclude with a short statement that your reason may cause such stereotypes of [BIASED-GROUP] in the context [EVENT].
\end{tcolorbox}
\end{center}

\section{More Experimental Results\label{app:moreresults}}
In this document, we present additional experimental results on several models, including \href{https://huggingface.co/microsoft/Phi-3.5-mini-instruct}{Phi-3.5-mini-instruct (3.8B)}, \href{https://huggingface.co/meta-llama/Meta-Llama-3-8B-Instruct}{Llama-3-8B-Instruct}, and \href{https://huggingface.co/mistralai/Mistral-7B-Instruct-v0.3}{Mistral-7B-Instruct-v0.3}. These results extend our analysis across diverse model architectures and parameter scales. Overall, the findings align with the conclusions reported in our main paper.

\subsection{Results of Phi-3.8b}
\begin{table*}[ht]
  \begin{center}
    
    \small
    \begin{tabular}{c c c c c c }
    \toprule

    \texttt{Phi3.5-mini} & \textbf{Age} & \textbf{Nation} & \textbf{Gender}& \textbf{SES}& \textbf{Disability} \\
    \midrule
    Baseline &.9176&1&.9492&.9985&.9803\\
    \midrule
    All &.9574&1&.9637&.9985&.9934\\
    \midrule
    All - Context &.9347&1&.9528&.9985&.9868\\
    All - Statement &\textbf{.9716}&\textbf{1}&\textbf{.9819}&\textbf{.9985}&\textbf{1}\\
    
      
    
    
   

      \bottomrule

    \end{tabular}
    \caption{Experimental Results on Phi-3.5-mini-instruct for \textbf{Self-correction} Across Different Discourse Constructions. \textbf{All} includes all possible components: context, statement, action. \textbf{All - *} indicates a component was removed from the setting of \textbf{All}. The optimal performance is highlighted in \textbf{bold}. 
    Across all experiments, \textbf{All - Statement} contributes to the optimal performance in all experiments, suggesting that LLMs do not need stereotype awareness for successful self-correction.}
    \label{tab:phi-main}
  \end{center}
\end{table*}

Table~\ref{tab:phi-main} presents the self-correction performance of Phi-3.5-mini-instruct after fine-tuning on three discourse constructions. All constructions demonstrate performance gains over the baseline, validating the effectiveness of our approach. Consistent with the findings in the paper, the discourse without the \texttt{Statement} component achieves the highest improvement, while removing \texttt{Context} results in a performance drop compared to the full construction (All), underscoring the critical role of contextual information.

Table~\ref{tab:phi-generalize} summarizes in-domain and out-of-domain generalization experiments for Phi-3.5-mini-instruct (3.8B). The results mirror those reported for LLaMA-3.2-3b-it in the paper, further demonstrate the robust out-of-domain generalization capabilities of LLMs of such sizes.

\begin{table*}[h]
  \begin{center}
    \small
    \begin{tabular}{c c c c c c | c c c }
    \toprule
    \texttt{Phi3.5-mini} &\textbf{Age}&\textbf{Nation}&\textbf{Gender}&\textbf{SES}&\textbf{Disability} &\textbf{SexOrientation} & \textbf{Physical} & \textbf{Religion}\\
    \midrule
    Baseline&0.9176&1&0.9492&0.9985&0.9803&0.995&0.970&0.943\\
     \texttt{Mixed} &0.9716&1&0.9691&0.9985&0.9934&0.9954&0.9824&0.9683\\
      \bottomrule
    \end{tabular}
    \caption{\small Self-correction In-domain and Out-of-domain Generalization of Phi-3.5-mini-instruct By Mixing the Fine-tuning Corpus of 5 Representative Stereotypes. Phi-3.5-mini-instruct(3.8b) shows both good in-domain and out-of-domain generalization}
    \label{tab:phi-generalize}
  \end{center}
\end{table*}

\subsection{Results of larger models}

\begin{table*}[ht]
  \begin{center}
    \small
    \begin{tabular}{c c c c c c | c c c c c }
    \toprule

      \texttt{Llama3|Mistral} & \textbf{Age} & \textbf{Nation} & \textbf{Gender}& \textbf{SES}& \textbf{Disability}& \textbf{Age} & \textbf{Nation} & \textbf{Gender}& \textbf{SES}& \textbf{Disability} \\

      \midrule
      Baseline &.906&.987&.955&.897&.993&.838&.837&.695&.810&.849\\
      \midrule

    All &.974&1.0&1.0&.964&\textbf{1.0}&.878&\textbf{.887}&.871&.955&.882\\
    \midrule
    All - Context &.926&.983&.982&.941&1.0&.869&.880&.742&.821&.862\\
    All - Statement &.983&\textbf{1.0}&\textbf{1.0}&\textbf{1.0}&.938&\textbf{.938}&.857&.860&\textbf{.988}&.882\\

    All - Action &.957&.997&1.0&958&1.0&.847&.833&.790&.929&.875\\
    Action &.940&.990&.969&.927&1.0&.889&.843&.735&.830&.862\\
    
    \makecell{All-Statement\\-subaction} &\textbf{.986}&.993&1.0&.973&1.0&.898&.873&\textbf{.877}&.939&.875\\

      \bottomrule

    \end{tabular}
    \caption{Experimental Results on Llama3-8b-it/Mistral-v0.3-7b-it for \textbf{Self-correction} Across Different Discourse Constructions. \textbf{All} includes all possible components: context, statement, action. \textbf{All - *} indicates a component was removed from the setting of \textbf{All}. The optimal performance is highlighted in \textbf{bold}. 
    Across all experiments, only three out of ten \textbf{All - Statement} (\textbf{Disability} in Llama, \textbf{Age} and \textbf{Nation} in Mistral) do harm to the performance compared to the \textbf{All}, suggesting that LLMs do not need stereotype awareness for successful self-correction.}
    \label{tab:llama8bAndMistral-main}
  \end{center}
\end{table*}
\begin{table*}[ht]
  \begin{center}
    \small
    \begin{tabular}{c c c c c c | c c c c c }
    \toprule

    \texttt{Llama3|Mistral} & \textbf{Age} & \textbf{Nation} & \textbf{Gender}& \textbf{SES}& \textbf{Disability}& \textbf{Age} & \textbf{Nation} & \textbf{Gender}& \textbf{SES}& \textbf{Disability}\\
    \midrule
    selfdiag baseline &.665&.540&.604&.777&.697&.645&.550&.682&.635&.618\\
    \textbf{selfcorr}$\rightarrow$ selfdiag  &\underline{.713}&\underline{.633}&\underline{.655}&\underline{.882}&\underline{.697}&\underline{.656}&\underline{.560}&.603&.600&\underline{.632}\\
      \bottomrule
    \end{tabular}
    \caption{\small Experimental Results of Llama3-8b-it/Mistral-v0.3-7b-it for Test Performance in Self-diagnosis Capability While Improving Self-correction. For all ten experiments, there are conflicts for only two of them.}
    \label{tab:llamaAndMistral-conflicts}
  \end{center}
\end{table*}

\begin{table*}[ht]
  \begin{center}
    
    \small
    \begin{tabular}{c c c c c c | c c c c c }
    \toprule

    \texttt{Llama3|Mistral} & \textbf{Age} & \textbf{Nation} & \textbf{Gender}& \textbf{SES}& \textbf{Disable}& \textbf{Age} & \textbf{Nation} & \textbf{Gender}& \textbf{SES}& \textbf{Disable}\\
    \midrule
    Baseline &.906&.987&.955&.897&.993&.838&.837&.695&.810&.849\\

    \texttt{Individual}&.983&1.0&1.0&1.0&.938&.938&.857&.860&.988&.882\\
     \texttt{Mixed} &.992&.997&.996&.969&.993&.966&.980&.964&1.0&.967\\
      \bottomrule

    \end{tabular}
    \caption{\small In-domain Generalization of Llama3-8b-it/Mistral-v0.3-7b-it By Mixing the Fine-tuning Corpus of Five Representative Stereotypes. \texttt{Individual} represents fine-tuning with the discourse within one stereotype. \texttt{Mixed} means that we mix the fine-tuning dataset of five stereotypes and test the fine-tuned model on each stereotype. The 7b/8b models shows great in-domain generalization capability.}
    \label{tab:llama3AndMistal-indomain}
  \end{center}
\end{table*}
This section presents experiments on larger models, LLaMA-3-8B-Instruct and Mistral-7B-Instruct-v0.3, using LoRA fine-tuning with a rank of 64 and a learning rate of 1e-5. 
It should be noted that we select Llama3-8b due to its architectural consistency with the 1B/3B models in our study. However, since the BBQ benchmark is widely used, this model has already been fine-tuned on it and achieves near perfect performance. Nevertheless, our framework still demonstrates measurable improvements.
Table~\ref{tab:llama8bAndMistral-main} summarizes the self-correction performance across all discourse constructions. Compared to smaller models, the 7B/8B parameter models achieve superior performance, with LLaMA-3-8B-Instruct attaining perfect accuracy in four out of five tasks. These results further validate the effectiveness of our proposed constructions for larger-scale LLMs.

Compared with smaller LLMs, there are some important observations. Removing the \texttt{Context} or \texttt{Action} components do harm to the performance relative to the full construction (\texttt{All}). However, only three out of ten \textbf{All - Statement} (specifically, \textit{Disability} in Llama, \textit{Age} and \textit{Nation} in Mistral) underperform the \textbf{All} construction. This suggests that that explicit stereotype awareness is \textit{not} essential for successful self-correction. These findings align with the shallow heuristics hypothesis presented in Section 4.2 of our paper.
Table~\ref{tab:llamaAndMistral-conflicts} present self-diagnosis performance when self-correction is enhanced. There are two key-observations: \textbf{(1)} The overall performance is not higher than smaller models significantly (primarily between 0.6-0.7). \textbf{(2)} The conflicts are slightly mitigated in larger-scale models. 
The empirical findings necessitate fine-grained analysis of self-diagnosis, which is essential to understand how LLMs perform moral tasks.


\end{document}